\documentclass[11pt]{article}

\usepackage[preprint]{acl}

\usepackage{times}
\usepackage{latexsym}

\usepackage[T1]{fontenc}

\usepackage[utf8]{inputenc}

\usepackage{microtype}

\usepackage{inconsolata}

\usepackage{graphicx}
\usepackage{amsmath}
\usepackage{subcaption}

\usepackage{float}

\usepackage{pdflscape}
\usepackage{tabularx}
\usepackage{lipsum}

\usepackage{longtable}

\usepackage{algorithm}
\usepackage{algpseudocode}

\usepackage{array} 
\usepackage{makecell} 
\usepackage[framemethod=tikz]{mdframed} 
\usepackage[export]{adjustbox} 
\usepackage{tikz}
\usepackage{booktabs}

\title{MusiChat: Vibe Composing for Music Creation}

\author{Callie C. Liao$^{*}$ \\
  Stanford University \\
  Stanford, CA \\
  \texttt{ccliao@cs.stanford.edu} \\\And
  Duoduo Liao$^{*}$ \\
  George Mason University \\
  Fairfax, VA \\
  \texttt{dliao2@gmu.edu} \\\And
  Ellie L. Zhang \\
  IntelliSky \\
  McLean, VA \\
  \texttt{elzhang@intellisky.org} \\}

\begin{document}
\maketitle
\def\thefootnote{*}\footnotetext{Equal contribution.}

\def\thefootnote{1}\footnotetext{MusiChat: \href{https://www.intellisky.org/share/musichat_demo.mp4}{demo video} and \href{https://musichat.intellisky.ai/}{AWS-based web tool}.}

\begin{abstract}

Recent advances in AI music generation have enabled users to create complete musical pieces from natural-language prompts. However, most existing systems follow a prompt-and-regenerate paradigm, making iterative refinement difficult because users must repeatedly recreate compositions instead of directly evolving existing musical ideas. We present MusiChat, a conversational vibe composing system that enables collaborative human--AI music creation through natural-language interaction and iterative refinement. At the core of MusiChat is a hierarchical controllable music generation framework that separates lyric-aligned musical structure generation from expressive surface realization, allowing flexible stylistic transformations and structure-preserving edits. The system integrates a large language model with a hybrid symbolic music engine through a memory-augmented architecture that maintains the active composition state and user history across interactions. A hybrid intent-routing mechanism further enables efficient interpretation of both precise musical edits and open-ended creative requests. Rather than regenerating compositions from scratch, MusiChat incrementally transforms an evolving musical artifact while preserving relevant musical structure and user intent. We evaluate MusiChat through objective analysis and human studies, achieving 95.31\% and 100\% accuracy for single- and multi-turn interactions, respectively, and obtaining like-to-dislike ratios of 2:1 for melody naturalness and 3:1 for musical quality. Our results demonstrate that MusiChat supports coherent multi-turn music authoring and interactive human--AI co-creation through a conversational interface.

\end{abstract}

\section{Introduction}

\begin{figure}[t]
  \centering
   \begin{mdframed}[linecolor=gray, linewidth=0.5pt, roundcorner=10pt]
  \centering
 \includegraphics[width=\textwidth]{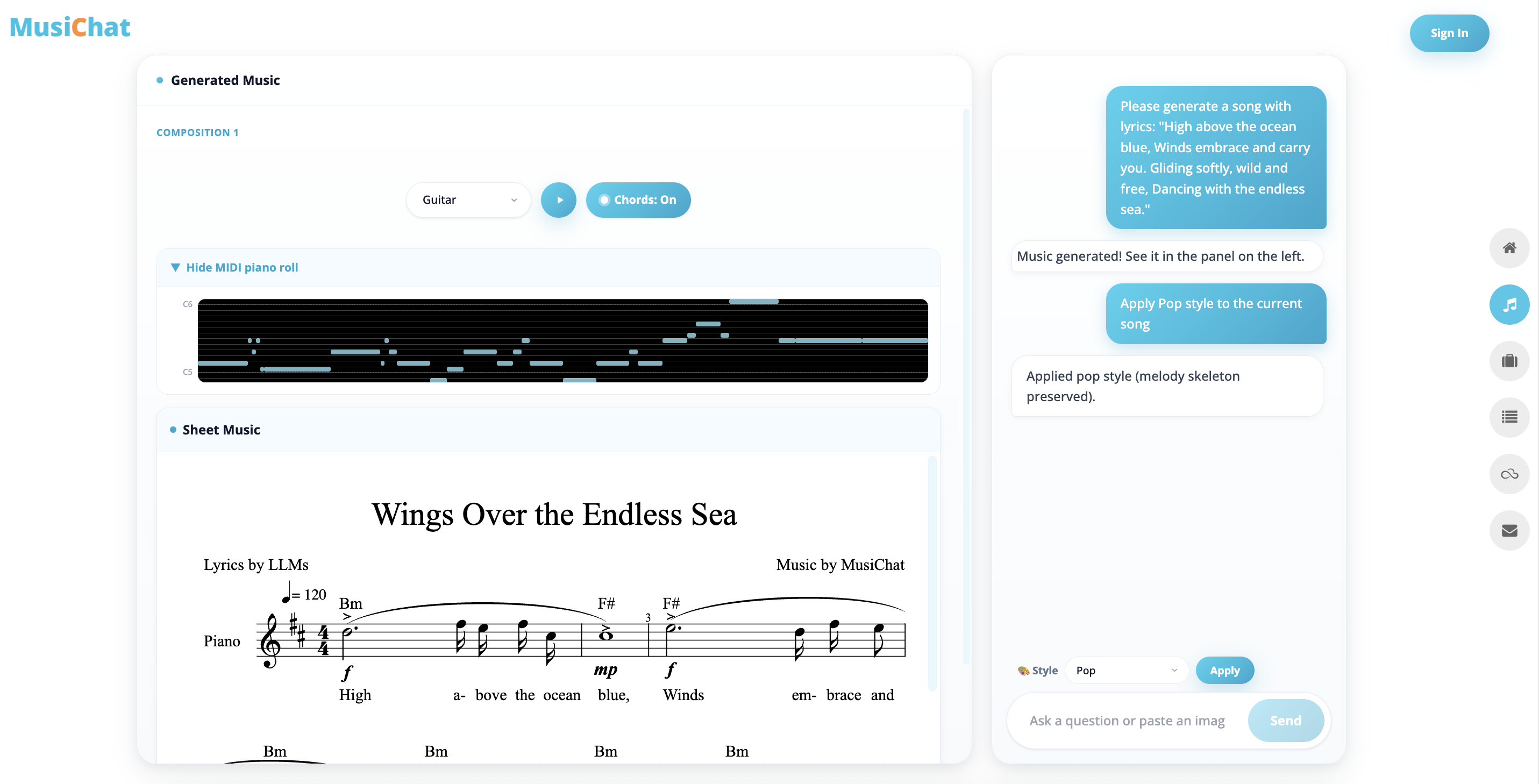}
 \end{mdframed}
  \caption{The MusiChat workspace. Left: engraved sheet music, MIDI player
  (instrument + chords controls), and piano roll. Right: the conversation and
  style selector.}
  \label{fig:ui}
\end{figure}

\begin{figure*}[ht!]
  \centering
  \includegraphics[width=\textwidth]{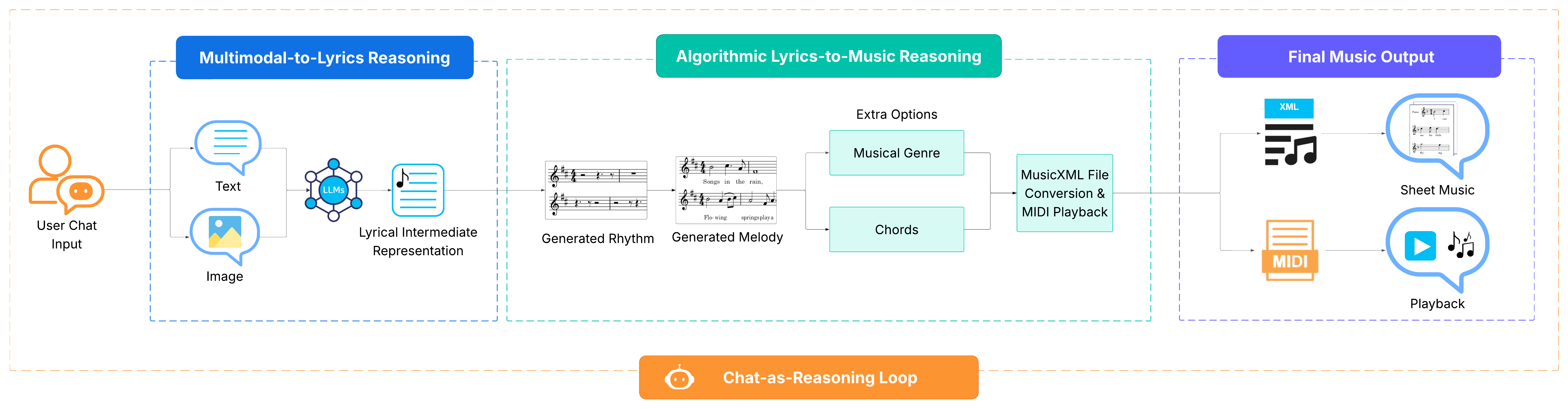}
  \caption{MusiChat framework for vibe composing for music co-creation. Multimodal inputs are first abstracted into an explicit lyrical representation via LLM-based reasoning. Lyrics serve as a persistent intermediate reasoning representation that guides hybrid symbolic music generation. A conversational interface enables iterative refinement by operating directly over these representations, transforming chat into an interactive reasoning environment.
}
  \label{fig:architecture}
\end{figure*}



With the proliferation of existing, well-known music generation models such as Suno~\cite{suno}, Google's Lyria~\cite{lyria}, Stable Audio 3.0~\cite{stableaudio3}, and MusicGen~\cite{musicgen}, music generation can now be accomplished with a single text, image, or audio prompt, making music creation possible for creators with little to no music experience. Some have also provided a wide variety of features to edit and refine the generated music, including allowing the artist to edit the music manually themselves within the product, adjusting the parameters as the music generates, and regenerating selected segments of the song through lyric editing or by just normal regeneration.

While they may offer segment regeneration, changes in music style as the music generates, and manual editing, a wish to regenerate only a portion of a musical feature (such as a melody or instrumental part) that \textit{maintains} the musical continuity of the original song remains out of scope, as users frequently receive a drastically different regeneration of that portion instead. Asking music generation models to revise a portion of the song is similar to asking state-of-the-art Large Language Models (LLMs) such as ChatGPT and Claude to modify only a paragraph of an essay while keeping others the same---in-turn refinement is currently not possible, as they only provide one-shot generation~\citep{huang2018musictransformer, payne2019musenet, dhariwal2020jukeboxgenerativemodelmusic, 10.1609/aaai.v39i24.34766}, where semantic content, structure, and surface realization are entangled in opaque representations, limiting interpretability, controllability, and collaborative use \citep{agostinelli2023musiclmgeneratingmusictext, wu2023melodyglmmultitaskpretrainingsymbolic, zhou2024llmsreasonmusicevaluation}. Symbolic models like MelodyGLM~\citep{wu2023melodyglmmultitaskpretrainingsymbolic} and MeloTrans~\citep{wang2024melotranstextsymbolicmusic} offer more transparency but use single-pass pipelines, while SongComposer~\citep{ding-etal-2025-songcomposer} advances lyric-aware generation, yet all remain largely data-driven and end-to-end.

Some alternatives to this end-to-end method would be to ask for a complete regeneration of the entire song but specify a slight change to only a portion of the song; however, that would likely result in a song with little to no similarity compared to the original, and due to a focus on expressing high-level intent, many music generation models currently do not have the ability to exactly pinpoint the area to make slight modifications to anyway. 

Under such limitations, artists are limited to either many  multi-turn regenerations using conversation as a medium~\citep{mdpi2025texttomusic, laban2025llmslostmultiturnconversation, zhang2024multimodalchainofthoughtreasoninglanguage, 10.1145/3771090} until they reach a somewhat coherent, satisfying result or editing the music in another software themselves. The former does not address the direct problem, and the latter requires a higher music production experience, confining many aspiring musicians. 

Therefore, we introduce MusiChat, a \textit{vibe composing} symbolic natural-language-to-music generation tool that provides reliable melodic, rhythmic, and lyrical in-turn refinement preserving all other parts of the song while producing musically coherent, intent-aware segments and retaining the musical structure. Similar to the term \textit{vibe coding}\footnote{https://x.com/karpathy/status/1886192184808149383}, the idea that a person can build software
by conversing with an LLM rather than writing code
directly, we introduce the term \textit{vibe composing} for music, where any melodic, rhythmic, and lyrical modifications to the music at any scope can be made purely using natural language. Contrary to current music generation methods, we incorporate a mix of hybrid symbolic music generation models to generate the desired music. By particularly including a deterministic design, in-turn refinement becomes feasible.    

In our demonstration, users not only have the ability to make basic modifications to features such as the author and title, but also to selected measures and lyrical segments through specific regeneration requests communicated with natural language. In addition to the listed features, users can alter the song style while preserving the song skeleton, hence allowing general refinement as well. With MusiChat, users are able to prompt, create, and refine musically, rhythmically, and lyrically without needing extensive musical experience, extending beyond the previous limitation placed upon precise musical and rhythmic modification requests for existing music generation models.


Our contributions are multifold:

\begin{itemize}
    \item \emph{MusiChat: A conversational vibe composing system} and an LLM-powered system that illustrates how conversational AI can facilitate collaborative human--AI music creation through continuous dialogue and iterative refinement.
    
    \item \emph{A hierarchical controllable music generation framework} that forms the foundation of vibe composition by separating lyric-aligned musical structure generation from expressive surface realization, enabling flexible, structure-preserving musical transformations.
    
    \item \emph{A comprehensive system evaluation} that evaluates interaction, musical structures, expressive variation, and perceptual quality through objective analysis and human studies.
    
\end{itemize}

We demonstrate MusiChat through representative composition scenarios, showing how conversational interaction, persistent memory, and incremental editing support a natural and iterative workflow for vibe composing with AI-assisted music generation.

\section{System Architecture}
\label{sec:system}

\subsection{System Overview}
\label{sec:overview}

MusiChat is a two-panel web-based workspace (Figure~\ref{fig:ui}) designed for vibe composing through conversational music creation and iterative refinement. The left panel presents the generated composition through engraved sheet music, a MIDI player with instrument and chord controls, and a collapsible piano roll for inspecting musical structure. The right panel provides the conversational interface for creating, modifying, and refining compositions through natural-language requests, exploring musical ideas, and applying stylistic transformations. 
A set of buttons on the far right side provides access to additional functions, including sign-in, composition history, input examples, and other utilities that support managing and revisiting creative work.

Figure~\ref{fig:architecture} illustrates the end-to-end system architecture. User requests are sent through an interface to a serverless backend, which performs intent routing, invokes foundation models for tasks such as lyric generation, and calls an in-process neuro-symbolic hybrid music engine for composition generation and transformation. Generated MusicXML and MIDI artifacts are then delivered to the client for notation rendering and audio playback.

The system maintains two complementary forms of state to support iterative vibe composing. \emph{Short-term composition state} captures the active piece across conversational turns, while \emph{persistent composition history} stores previous compositions, prompts, and user ratings for authenticated users in a document database, enabling users to revisit and continue prior creative explorations.



 \begin{figure*}
    \centering
    \includegraphics[width=0.9\linewidth]{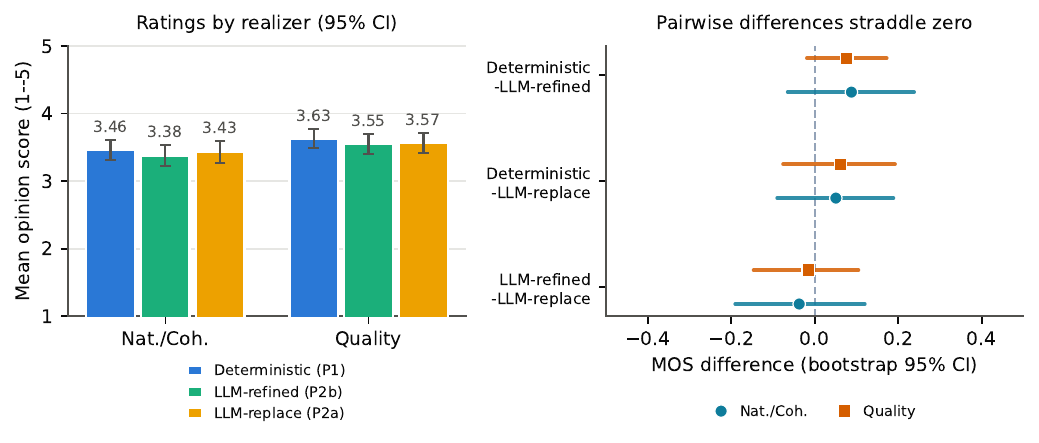}
    \caption{\emph{Left:} mean opinion
  score with 95\% Confidence Intervals (CIs) per realizer. \emph{Right:} percentile-bootstrap 95\% CIs for every
  pairwise per-participant Mean Opinion Score (MOS) difference.}
    \label{fig:ratings}
  \end{figure*}

\subsection{Natural Language Interaction}
\label{sec:nli}

The conversational loop forms the core interaction paradigm of vibe composing. Users create and refine compositions through natural-language instructions or image inputs, which are interpreted and mapped to structured musical operations. Requests are categorized into generation, parameter modification (e.g., key, tempo, and time signature), symbolic editing (e.g., title or pitch changes), style transformation, musical analysis, or general conversation. This intent-driven workflow enables iterative refinement while maintaining user control over both structural and expressive aspects of the composition.

\paragraph{Hybrid intent routing.}
MusiChat employs a hybrid intent-routing architecture that combines deterministic rules with LLM-based reasoning. Lightweight \emph{deterministic detectors} using regular expressions and keyword matching handle frequent, unambiguous edits (e.g. ``change to C major'' or ``make it faster'') without LLM calls. For open-ended or ambiguous requests, an \emph{LLM-based router} classifies intent and delegates tasks to the appropriate agent, including music generation, key, time signature, tempo, analysis, or general conversation. This hybrid design reduces latency and cost for structured edits while retaining the flexibility needed for natural-language vibe composing.

\paragraph{Multimodal lyric generation.}
The system supports multimodal lyric generation by allowing users to initiate composition from either textual or visual inputs. When provided with text, an LLM generates short lyrics that serve as the semantic foundation for music generation. When users provide an \emph{image}, a multimodal model interprets the visual content and generates image-inspired lyrics, supporting visual experiences such as photographs to be transformed into musical compositions. The resulting lyrics are hierarchically aligned with the generated melody at the levels of sections, phrases, words, and syllables, providing the foundation for subsequent lyric-aware composition and editing.

\paragraph{Persistent composition state.}
Edits operate on the \emph{current} composition rather than as independent generation requests. The backend maintains the working song state, while the client re-sends relevant context, including the previous prompt, composition metadata, and pending suggestions across interactions. This design ensures iterative refinement despite stateless serverless execution, allowing each conversational turn to transform the same musical artifact. A ``regenerate'' request rebuilds the composition from the initial generation process, whereas an in-place edit updates the existing piece and replaces the current composition card rather than creating a new version.


\subsection{Hybrid Symbolic Music Engine}
\label{sec:engine}

Music generation is performed by a self-contained Python engine integrated into the backend, without relying on an external model server. The engine adopts a \emph{hierarchical representation} that forms the foundation of vibe composition. It first derives a stable \emph{musical skeleton} from the input lyrics and then applies a pluggable \emph{surface realizer} to produce style-specific musical realizations while keeping the underlying skeleton unchanged.

\paragraph{Lyric-to-music skeleton generation.}
Lyrics provide the semantic and prosodic foundation for melody generation. The engine analyzes linguistic \emph{stress} patterns to infer an appropriate time signature and hierarchically segments lyrics into sections, phrases, words, and syllables~\cite{Liao_2023}. Melodic notes are assigned to syllables, with rests inserted at phrase boundaries to reinforce musical structure. Pitch selection follows a buffer-constrained, predominantly stepwise traversal across measures, with smoothing at phrase transitions to promote melodic continuity. The resulting melody, rhythm, phrase structure, and lyric alignment form the compositional backbone preserved throughout subsequent editing operations~\cite{CLiao2024, MusicAIR2025, Liao2022}.

\paragraph{Surface realization (three realizers).}

A post-generation styling subsystem transforms the neutral compositional scaffold into style-specific renditions through a fixed pipeline of rhythm, harmony, melody, dynamics, and phrasing transformations. Parameterized by specifications for eight genres (\emph{Classical, Pop, Jazz, Rock, Waltz, Lo-fi, Cinematic}, and \emph{R\&B}), the subsystem decouples style realization from the underlying skelton, supporting flexible vibe composing while preserving melodic continuity, rhythmic structure, and lyric alignment.

The framework provides three interchangeable realizers over the same scaffold: (i) a \emph{deterministic} realizer applying rule-based transformations; (ii) an \emph{LLM-refined} realizer enriching the deterministic output with ornaments, chromatic embellishments, and articulation; and (iii) an \emph{LLM-driven style-replacement} realizer generating surface-level style realization through the language model. All realizers preserve the melodic scaffold and note onsets while modifying surface attributes such as pitch realization, harmony, articulation, dynamics, and expression. Each styling operation starts from a fresh copy of the base composition to prevent accumulated transformations.

\paragraph{Chord harmonization.}

During generation, the system assigns chord symbols to the notation and produces a harmonized melody with chord accompaniment. Harmony is modeled as an independent layer over the musical skeleton, enabling flexible vibe composing while preserving the melody. Users can control chord realization through interface controls or natural-language commands, with the \emph{Chords} toggle managing playback and chord-symbol visibility.

\paragraph{Multi-instrument playback and robustness.}

The playback system supports ten instruments (e.g., \emph{Grand Piano, Electric Piano, Guitar, Flute, Violin, Saxophone, Cello, Clarinet,} and \emph{Drums}) through client-side soundfont rendering, enabling diverse sonic exploration during vibe composing. To maintain a robust creative workflow, the generation pipeline uses graceful degradation through a fallback hierarchy: styled composition with chords, chords-only output, and finally, melody-only output. This ensures that transformation failures reduce expressive richness rather than preventing playable results, preserving iterative music creation.

  

\section{Demonstration and Experiments}
\label{sec:eval}

We conducted a mix of quantitative feature editing and human evaluation to evaluate this paradigm mainly on its consistency, ability to retain the correct features over multiple turns, the naturalness or melodic consistency of the generated music, and the overall musical quality. The former two qualities are evaluated through feature editing, and the latter two are evaluated through our conducted human study. More detailed evaluation for objective, music-theory-based metrics have been conducted in \cite{MusicAIR2025}.


 


\paragraph{Experimental Setup.}
To evaluate the proposed MusiChat paradigm, we developed a prototype web tool that facilitates both image-to-lyrics and text-to-lyrics generation. The MusiChat prototype is deployed both locally and on AWS, and uses the Amazon Nova 2 multimodal AI model for LLM-based lyric generation from images and text description as well as for general agentic reasoning.

\begin{table*}[h]
\centering
\footnotesize
\setlength{\tabcolsep}{2.5pt}
\renewcommand{\arraystretch}{1.1}

\caption{Interactive Feature Editing Performance. (a) Single-Turn. (b) Multi-Turn.}

\begin{minipage}[t]{0.41\textwidth}
\centering
\textbf{(a) Single-Turn}\\
\vspace{1mm}

\begin{tabular}{lcc lcc}
\toprule
Feature & H & L & Feature & H & L \\
\midrule
Title   & 8 & 0 & Title+KS              & 8 & 0 \\
KS      & 8 & 0 & KS+Tempo             & 8 & 0 \\
Tempo   & 8 & 0 & KS+Tempo+Pitch       & 7 & 1 \\
Pitch   & 6 & 2 & Tempo+Pitch+Title    & 8 & 0 \\
\midrule
TOTAL   & 30 & 2 &                      & 31 & 1 \\
\midrule
\multicolumn{6}{c}{Accuracy: 95.31\%} \\
\bottomrule
\end{tabular}
\end{minipage}
\hfill
%
\begin{minipage}[t]{0.57\textwidth}
\centering
\textbf{(b) Multi-Turn}\\
\vspace{1mm}

\begin{tabular}{lcc lcc}
\toprule
Feature & H & L & Feature & H & L \\
\midrule
Title,KS        & 8 & 0 & KS+Pitch,Tempo   & 5 & 0 \\
KS,Tempo        & 8 & 0 & KS,Tempo+Pitch   & 5 & 0 \\
KS,Tempo,Pitch  & 5 & 0 &                  &   &   \\
Pitch,KS        & 5 & 0 &                  &   &   \\
\midrule
TOTAL           & 26 & 0 &                  & 10 & 0 \\
\midrule
\multicolumn{6}{c}{Accuracy: 100\%} \\
\bottomrule
\end{tabular}
\end{minipage}

\vspace{2mm}

\centering
\footnotesize
\textbf{Overall Accuracy: 97\% (H=97, L=3)}

\vspace{2mm}
\centering
\parbox{0.95\textwidth}{
\footnotesize
\centering
\textit{H: Hit, L: Loss, KS: Key Signature.}
}
\label{tab:musichat_eval_results}
\end{table*}

\subsection{Feature Editing Evaluation}
We evaluated feature editing performance across single-turn and multi-turn settings, each further divided into single- and multi-feature edits, where each turn contains one or more editing requests. In total, we collected 64 single-turn and 36 multi-turn edits involving musical attributes such as pitch, tempo, and key signature, either individually or in combination. For instance, the "KS+Pitch, Tempo'' category featured two turns: one turn for a key signature and pitch change, and a separate turn for a tempo change. For each feature, the number of hits and losses was recorded, with hits defined as having all correct editing and losses defined as having one or more incorrect changes. Detailed results are reported in Table \ref{tab:musichat_eval_results}. Across these settings, the system consistently demonstrates strong performance. In the single-turn case, it achieves an accuracy of 95.31\%, with minor errors primarily in pitch-related configurations. In the multi-turn setting, it attains 100\% accuracy, indicating that iterative editing does not introduce error accumulation and remains stable under sequential or compositional edits. Overall, the system achieves 97\% accuracy, highlighting its robustness and scalability across both single- and multi-feature editing scenarios.

\subsection{Human Evaluation}

\subsubsection{Study Design}

The survey contained 35 participants with self-reported musical experience: 18 reported some musical training, 9 no formal training, and 8 were professional musicians or had extensive musical experience. Participants were not informed about how the excerpts were produced to avoid algorithm-aversion bias.

Each participant was asked to evaluate the musical coherence and quality of each short excerpt from 1 to 5, with 1 being incoherent or low quality and 5 being very coherent and high quality. In each section, participants listened to three melodies presented in randomized order from the same set of lyrics. Each melody was generated with a randomized musical genre, and it was a distinct realization from the three categories: deterministic, LLM-refined, and LLM-replaced. The specific information regarding each melody was not disclosed to the participants, each melody being labeled as A, B, and C to maintain objectivity. 

\subsubsection{Results}

From the left figure of Figure \ref{fig:ratings}, the Mean Opinion Scores (MOS) with 95\% Confidence Intervals (CI) for all three realizers are around average, but slightly on the higher end. The highest mean opinion score among the three realizers is the deterministic realizer. Looking at the figure on the right, the pairwise differences between the realizers are all around zero, so there is no significant difference between the three realizers for their mean opinion scores; however, the deterministic realizer overall has higher pairwise differences compared to the rest of the combinations.

Figure \ref{fig:raincloud} sorts the naturalness ratings by listener experience and the realizer. From a distribution perspective, the group with zero musical experience and the group with extensive musical experience had distributions that skewed left, whereas the group with some musical experience did not particularly concentrate at any point. Notably, the mean is the highest for the group with zero musical experience and the lowest for the group with some musical experience, and the mean for all three groups concentrated in the region between 3 and 4, indicating that at best, the melodies were somewhat natural, and at worst, the listeners felt indifferent to the naturalness. Based on this figure and values from Table \ref{tab:byexp}, within each group, the deterministic realizer (3.26) had the highest average ratings for listeners with some experience; LLM-refined (3.90) had the highest average for listeners with no experience; and LLM-replace (3.46) had the highest average for listeners with extensive experience. 

Figure \ref{fig:acceptance} displays a pooling of ratings across all realizers, songs, and participants, with the only categorization being based on the survey questions. For naturalness ratings, approximately 50\% of the ratings were 4 or higher compared to the 25\% that were 2 or lower, giving an overall 2:1 ratio on naturalness. Similarly, the overall quality ratings observe a 55\% and 18\% like and dislike rate, giving an approximately 3:1 ratio on the category. With both questions heavily favoring the ratings of 4 or higher, it indicates that the listeners generally liked the generated music. 


\subsection{Conversational Evaluation}

Figure \ref{fig:conv} evaluates the consistency and the preservation of multiple features across multiple instances of regenerating and editing songs. On the top left, the figure demonstrates that composition consistency is 100\%, while intent fulfillment is the same with the exception of mood consistency. On the top right, the melody preservation for the MusiChat editor is at minimum more than two times the melody fidelity of a basic regenerating baseline, and at most more than three times the baseline. At the bottom left, while both methods achieved full lyric alignment, the MusiChat editor achieved 2-4 times better preservation than the baseline. At the bottom right, the figure reveals that despite having 100\% preservation for aspects that did not require change, the baseline tended to change the melody over time and experienced error accumulation, whereas MusiChat largely preserved it compared to the baseline.

Overall, results demonstrate coherent, perceptually realistic, and structurally consistent music generation under interactive use. As a result, the autonomous generation can provide immediate inspiration and support iterative refinement while preserving flexibility in aligning with user intent. Additionally, the combined approach of being unified, language-driven, and multimodal helps improve coherence and creative control.

\section{Conclusions}


We present MusiChat, a conversational vibe composing text-to-music generation system that utilizes a hybrid architecture of deterministic and music generation models. Due to this architecture, MusiChat supports in-turn iterative refinement for each desired revision that still semantically, melodically, and rhythmically preserves the untouched features, positioning natural language as the main driver for co-creation in creative AI. The workflow that MusiChat supports is what we would call \textit{vibe composing}. Our 35-person study on the naturalness and overall quality of the melody showed that the ratio of like to dislike was 2:1 for naturalness and 3:1 for overall quality, and MusiChat achieved 95.31\% and 100\% accuracies in single-turn and multi-turn refinements, respectively. By supporting a full vibe composing workflow, MusiChat can further foster human creativity by expanding the music creative process to users with little to no musical experience. In future work, we plan to scale the extent of music generation to multi-instrument or polyphonic compositions and expand to more modalities.

\section*{Broader Impact}

This work advances interpretable and collaborative creative AI by reframing music generation as a language-centric reasoning process rather than opaque synthesis, enabling inspection, editing, and human oversight that are difficult to achieve in end-to-end neural systems. The approach supports responsible creative AI by reducing reliance on black-box generative models for musical output while maintaining real-time interactivity. 

Potential risks include over-reliance on AI-generated lyrical interpretations, which may narrow creative exploration. We mitigate these risks by exposing all intermediate representations, supporting conversational revision, and allowing users to override or refine both lyrical interpretations and algorithmic composition decisions. Overall, the framework demonstrates how language-based reasoning combined with algorithmic realization enables transparent, scalable, and human-centered creative systems.

\section*{Limitations} 
Our framework centers natural language as an explicit reasoning representation, using lyrics and conversation to guide music generation. While this improves interpretability and control, it depends on the clarity and expressiveness of linguistic input, as ambiguous or underspecified lyrics may lead to under-constrained musical structure. Moreover, the multimodal-to-lyrics stage relies on LLMs to translate multimodal inputs into descriptive lyrical representations. Errors, biases, or omissions in multimodal–semantic interpretation may propagate into downstream reasoning. Furthermore, the melody generation in the lyrics-to-music pipeline is implemented using a purely algorithmic composition engine, ensuring determinism, interpretability, and real-time performance, but it may result in less flexibility for certain musical genres. Additionally, while the symbolic stage is computationally lightweight, upstream LLM-based reasoning may still reflect model biases.

\section*{Ethics Statement}

The lyrics-to-melody algorithm powering the music generation does not contain any copyright concerns since it is purely algorithmic. However, the chord generation, LLM-Replace and LLM-Refine  and multimodal feature extraction do contain the usage of LLMs. All photo examples presented in this paper and incorporated into the MusiChat web tool were taken by the authors and are free from copyright restrictions.

\bibliography{custom}

\newpage
\appendix

\newpage
\clearpage
\onecolumn

\appendix

\section*{Appendix}
\label{sec:appendix}

\subsection* {Demo Video and Live System}
\begin{itemize} 
    \item Explore \href{https://musichat.intellisky.ai/}{the AWS-based web tool} to interact with MusiChat to generate music online, analyze and edit generated sheet music, and play the music in the conversational interface in real time. 
    \item Watch the  \href{https://www.intellisky.org/share/musichat_demo.mp4}{MusiChat demo video} featuring interactive demonstrations of human-AI music co-creation, including simple or complex queries, editing, analysis, playback, etc.
\end{itemize}

\begin{table}[h]
\centering
\small
\begin{tabular}{p{2.8cm}p{0.8cm}p{1.cm}p{1.cm}}
\toprule
\textbf{Naturalness MOS} & \textbf{Det.\ (P1)} & \textbf{Refine (P2b)} & \textbf{Replace (P2a)} \\
\midrule
No training \; (n=9)   & 3.86 & \textbf{3.90} & 3.83 \\
Some training (n=18)   & \textbf{3.26} & 3.17 & 3.20 \\
Professional \, (n=8)  & 3.43 & 3.22 & \textbf{3.46} \\
\bottomrule
\end{tabular}
\caption{Naturalness MOS by musical experience. Non-experts (target users) give the highest ratings, while realizers remain comparable across groups.}
\label{tab:byexp}
\end{table}

  \begin{figure*}[h]
    \centering
    \includegraphics[width=\linewidth]{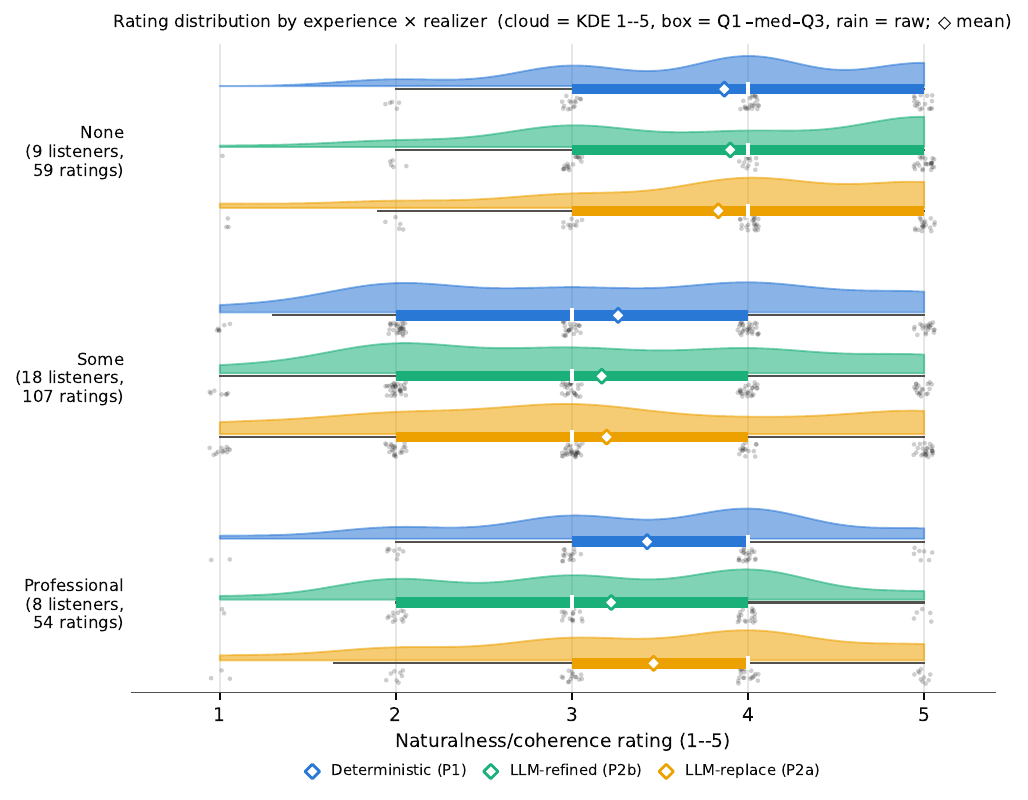}
    \caption{Raincloud of naturalness ratings by listener experience $\times$ realizer.}
    \label{fig:raincloud}
  \end{figure*}

  \begin{figure*}[h]
    \centering
    \includegraphics[width=1\linewidth]{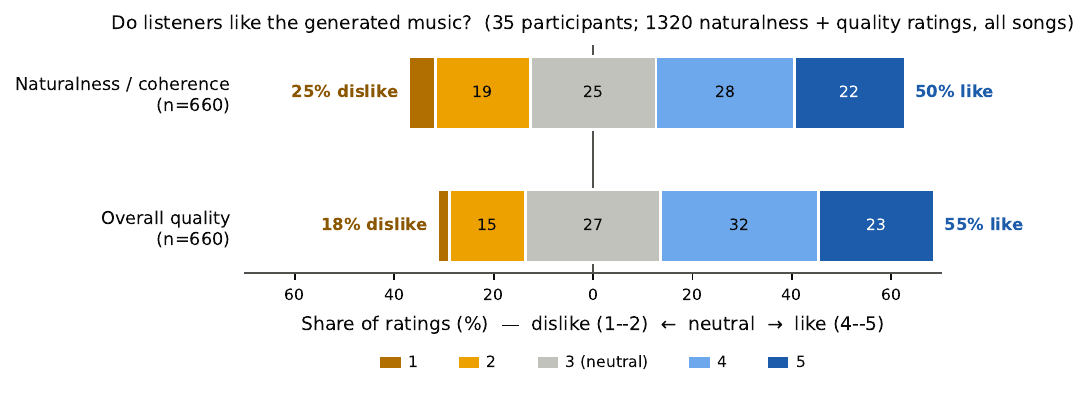}
    \caption{Pooling of ratings across all realizers, songs, and participants.}
    \label{fig:acceptance}
  \end{figure*}

  \begin{figure*}[h]
    \centering
    \includegraphics[width=\linewidth]{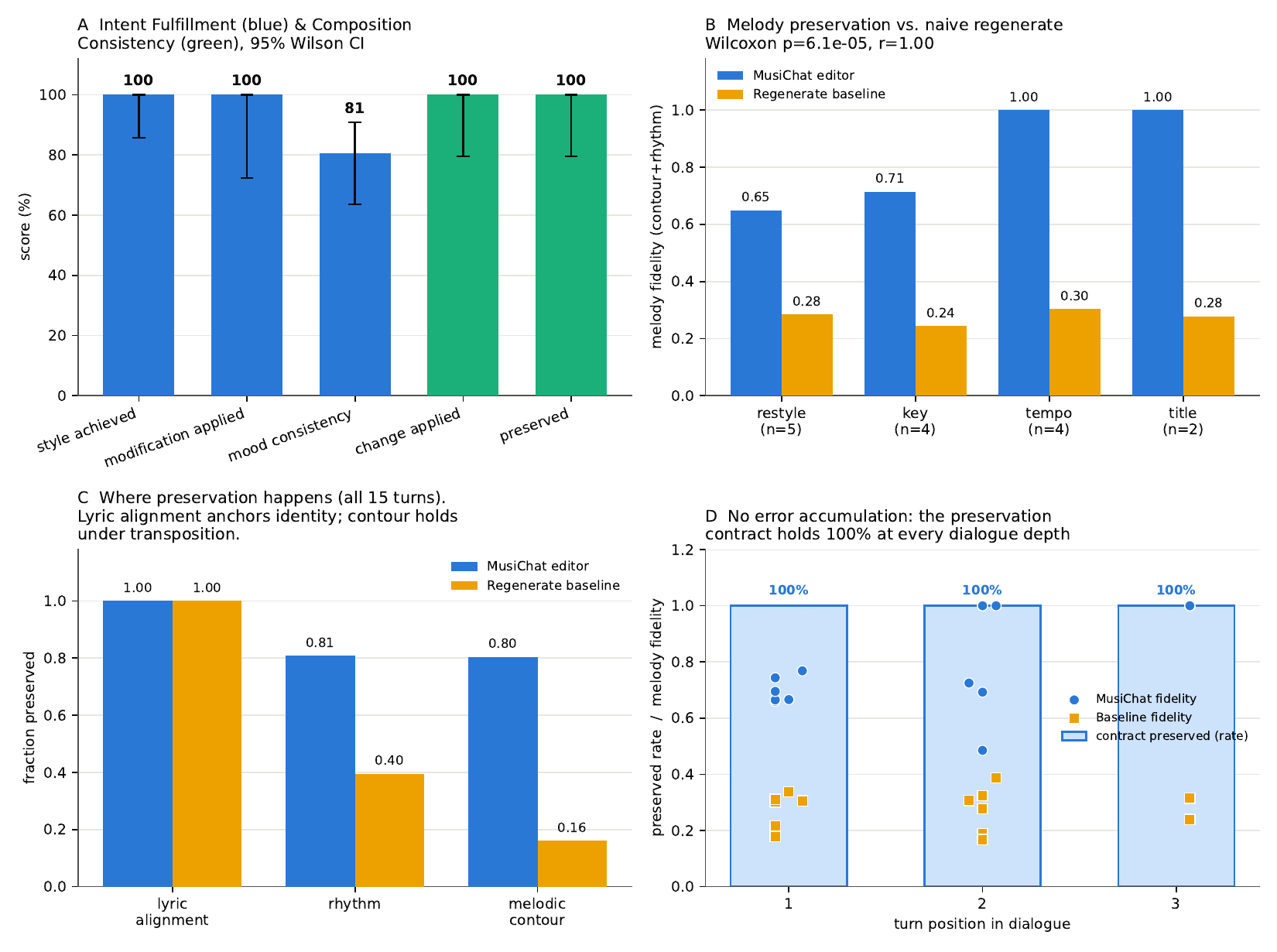}
    \caption{Conversational evaluation. \textbf{(A)}~The two headline metrics with
Wilson 95\% confidence intervals (CIs). \textbf{(B)}~MusiChat melody preservation vs.\ the regenerate-from-scratch baseline, per edit type. \textbf{(C)}~Preservation by feature. \textbf{(D)}~MusiChat and baseline fidelities versus the rate of contract preserved.}
    \label{fig:conv}
  \end{figure*}


\end{document}